\documentclass[10pt,twocolumn,letterpaper]{article}

\usepackage{iccv}
\usepackage{times}
\usepackage{epsfig}
\usepackage{graphicx}
\usepackage{amsmath}
\usepackage{amssymb}
\usepackage{comment}
\usepackage{booktabs}
\usepackage{subfig}
\usepackage{placeins}
\usepackage{textpos}

\usepackage[pagebackref=true,breaklinks=true,letterpaper=true,colorlinks,bookmarks=false]{hyperref}

\iccvfinalcopy 

\def\iccvPaperID{19} 

\ificcvfinal\fi

\begin{document}

\newcommand{\subfigureautorefname}{\figureautorefname}

\title{Evaluation of Deep Learning on an Abstract Image Classification
  Dataset}

\author{Sebastian Stabinger\\
  {\tt\small Sebastian.Stabinger@uibk.ac.at}
  \and
  Antonio Rodr\'iguez-S\'anchez\\
  {\tt\small Antonio.Rodriguez-Sanchez@uibk.ac.at}
  \and
  University of Innsbruck\\
  Technikerstrasse 21a, Innsbruck, Austria\\
}

\maketitle
\thispagestyle{empty}

\begin{textblock*}{200mm}(.2\textwidth,-7cm)
  Copyright IEEE. To be published at proceedings of MBCC/ICCV2017
\end{textblock*}

\begin{abstract}
  Convolutional Neural Networks have become state of the art methods
  for image classification over the last couple of years. By now they
  perform better than human subjects on many of the image
  classification datasets. Most of these datasets are based on the
  notion of concrete classes (\ie images are classified by the type of
  object in the image). In this paper we present a novel image
  classification dataset, using abstract classes, which should be easy
  to solve for humans, but variations of it are challenging for CNNs.
  The classification performance of popular CNN architectures is
  evaluated on this dataset and variations of the dataset that might
  be interesting for further research are identified.
\end{abstract}

\section{Introduction}
Convolutional Neural Networks have become the method of choice for
image classification since the system by Krizhevsky \etal
\cite{krizhevsky2012imagenet} won the ImageNet competition in 2012 by
a large margin. In 2015 Russakovsky \etal
\cite{russakovsky2015imagenet} reported the classification accuracy of
human subjects, on the same dataset, to be around 94.9\% correctly
classified images. In the same year, He \etal \cite{he2015delving}
were the first to present a network that outperformed human subjects
on ImageNet. Since then, image classification is often perceived as
either solved, or in the process of being solved.

Popular datasets used for image classification like
MNIST\cite{lecun1998gradient}, ImageNet\cite{russakovsky2015imagenet},
PASCAL\cite{Everingham10}, and
CIFAR10/100\cite{krizhevsky2009learning} all classify the images by
the type of a prominent object or feature in the image. We will call
such classes \emph{concrete classes}. Concrete classes have in common
that they can be identified by analyzing local features, or the
distribution of multiple local features. In this paper, we present a
dataset that consists of \emph{abstract classes}. Abstract classes
imply that images can not be classified by simply considering local
features. In our case, the two types of classes are
identity/non--identity and symmetry/non--symmetry.

\begin{figure}[t]
  \centering
  \subfloat[Identity task]
  {\includegraphics[width=.23\textwidth]{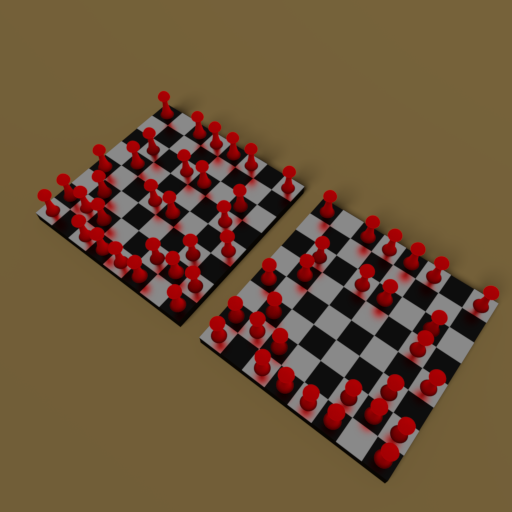}}
  \subfloat[Symmetry task]
  {\includegraphics[width=.23\textwidth]{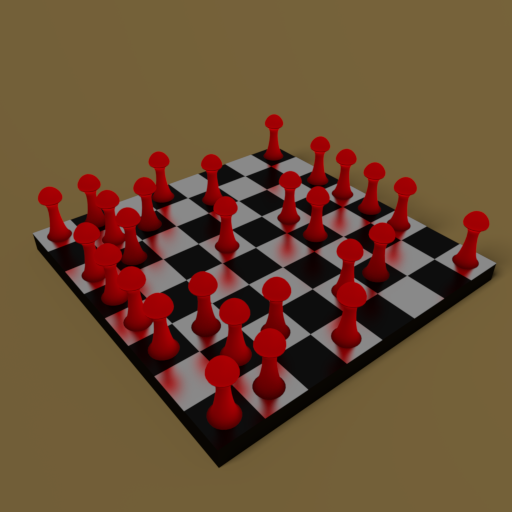}}
  
  \caption{Example images of the two classification tasks in the
    dataset. In both cases two of the pawns are out of place.}
  \label{fig:intro}
\end{figure}

\section{Related Work}

Fleuret \etal \cite{fleuret2011comparing} have already presented a
dataset with abstract classes, using very simple black and white line
drawings. This dataset is somewhat reminiscent of the ``Bongard
problems'', presented by Bongard in 1970 \cite{bongard1970} as a set
of problems that, according to Bongard, neural networks would never be
able to solve (though he did not have simple classification in mind,
but a textual description of what separates the two classes). In
previous work \cite{stabinger2016learning} \cite{stabinger201625}, we
have tested different convolutional neural network architectures on
the dataset by Fleuret \etal and came to the conclusion that current
CNN architectures have shortcomings when shape comparison is needed to
distinguish two classes. As Dodge \etal \cite{dodge2017study} argue,
the Fleuret dataset is too simplistic and too far from natural images
to draw any practical conclusions from it. Our goal is to present a
more realistic dataset, with abstract classes, that is equally hard to
classify for CNNs.

\section{Dataset}
\def\thumbwidth{.1\textwidth}
\def\imgspace{10mm}
\begin{figure*}[pt]
  \centering
  \subfloat[a][Fixed camera position.]{
    \includegraphics[width=\thumbwidth]{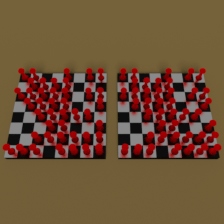}
    \includegraphics[width=\thumbwidth]{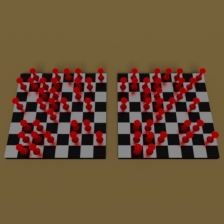}
    \includegraphics[width=\thumbwidth]{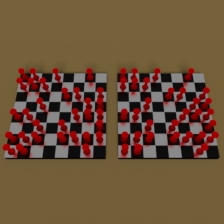}
    \hspace{\imgspace}
    \includegraphics[width=\thumbwidth]{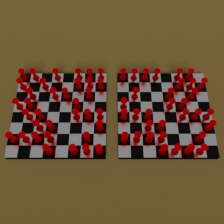}
    \includegraphics[width=\thumbwidth]{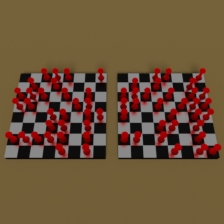}
    \includegraphics[width=\thumbwidth]{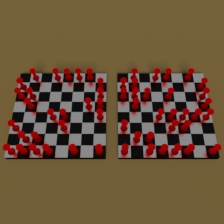}
    \label{fig:simfix}
  }
  
  \subfloat[Random camera translation.]{
    \includegraphics[width=\thumbwidth]{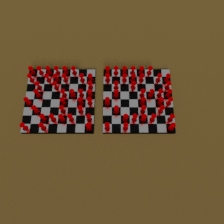}
    \includegraphics[width=\thumbwidth]{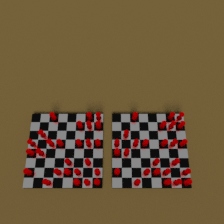}
    \includegraphics[width=\thumbwidth]{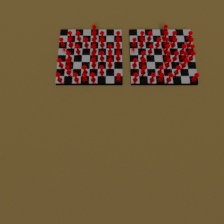}
    \hspace{\imgspace}
    \includegraphics[width=\thumbwidth]{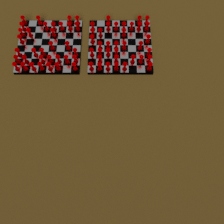}
    \includegraphics[width=\thumbwidth]{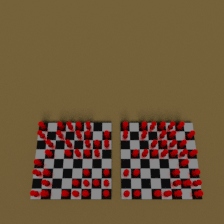}
    \includegraphics[width=\thumbwidth]{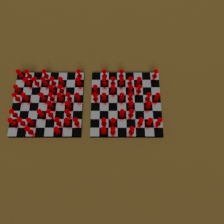}
    \label{fig:simtrans}
  }

  \subfloat[Random board position.]{
    \includegraphics[width=\thumbwidth]{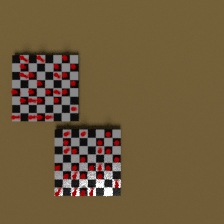}
    \includegraphics[width=\thumbwidth]{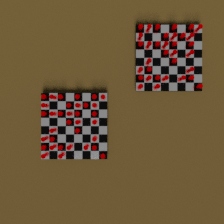}
    \includegraphics[width=\thumbwidth]{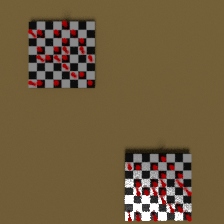}
    \hspace{\imgspace}
    \includegraphics[width=\thumbwidth]{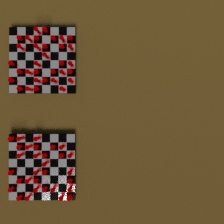}
    \includegraphics[width=\thumbwidth]{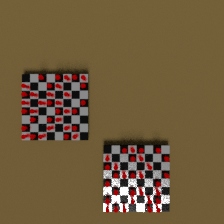}
    \includegraphics[width=\thumbwidth]{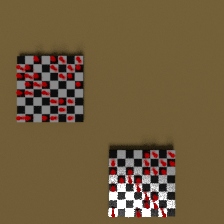}
    \label{fig:simrand}
  }

  \subfloat[Random camera position on sphere.]{
    \includegraphics[width=\thumbwidth]{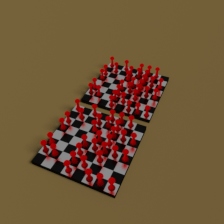}
    \includegraphics[width=\thumbwidth]{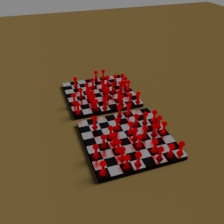}
    \includegraphics[width=\thumbwidth]{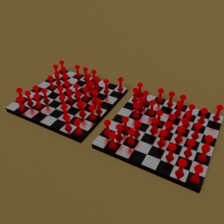}
    \hspace{\imgspace}
    \includegraphics[width=\thumbwidth]{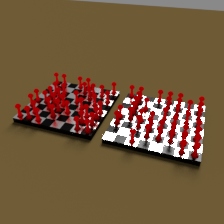}
    \includegraphics[width=\thumbwidth]{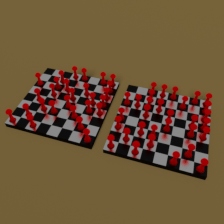}
    \includegraphics[width=\thumbwidth]{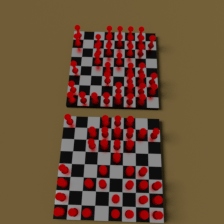}
    \label{fig:simrot}
  }
  \caption{Different, tested variations of the identity--task. The
    left group of images are samples from the identity class. The
    right group are from the non-identity class with ten pawns that
    are out of place.}
\end{figure*}

The presented dataset consists of two separate tasks:
\begin{enumerate}
\item The \emph{symmetry}--task: The system has to decide whether an
  arrangement of red pawns on a checkerboard is symmetric along one of
  the mid lines of the checker board, or not.
\item The \emph{identity}--task: The system has to decide whether
  the arrangement of red pawns on two checkerboards is identical, or
  not.
\end{enumerate}
There are multiple reasons for selecting these specific tasks:
\begin{enumerate}
\item The use of checkerboards, with randomly positioned pawns, allows us
  to very easily generate random samples, without inadvertently
  introducing additional, unwanted clues to the dataset. As we could
  show \cite{stabinger201625} for the dataset by Fleuret \etal
  \cite{fleuret2011comparing}, these unintended clues can be used by
  CNNs to classify images, and might lead to wrong conclusions about
  what CNNs are able to learn.
\item Although the images have a random component (the position of the
  pawns on the board), the images are still semi-realistic and show a
  simplified representation of what might be observable in reality.
\item According to the gestalt principles \cite{Todorovic:2008}, symmetry
  is an important property for humans to understand and order the
  world. It therefore seemed like a good choice for one of the tasks.
  The identity--task was chosen since the tests on the dataset by
  Fleuret \etal \cite{fleuret2011comparing} showed that CNNs have
  specific weakness when it comes to detecting identity.
\end{enumerate}

\begin{figure*}[t]
  \centering
  \subfloat[Fixed camera position.]{
    \includegraphics[width=\thumbwidth]{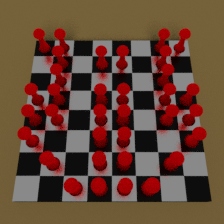}
    \includegraphics[width=\thumbwidth]{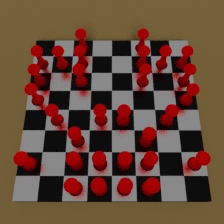}
    \includegraphics[width=\thumbwidth]{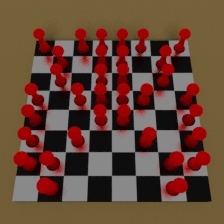}
    \hspace{\imgspace}
    \includegraphics[width=\thumbwidth]{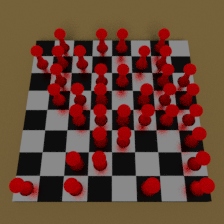}
    \includegraphics[width=\thumbwidth]{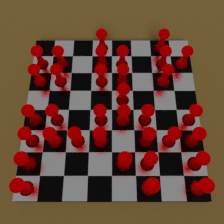}
    \includegraphics[width=\thumbwidth]{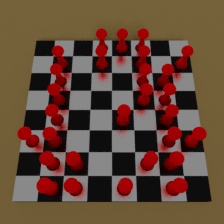}
    \label{fig:symfix}
  }

  \subfloat[Random camera translation.]{
    \includegraphics[width=\thumbwidth]{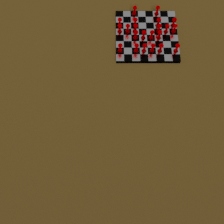}
    \includegraphics[width=\thumbwidth]{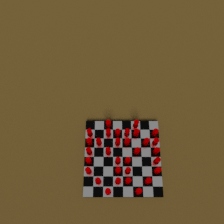}
    \includegraphics[width=\thumbwidth]{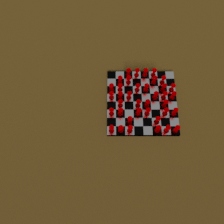}
    \hspace{\imgspace}
    \includegraphics[width=\thumbwidth]{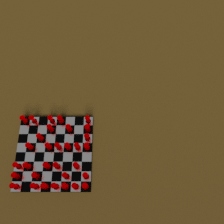}
    \includegraphics[width=\thumbwidth]{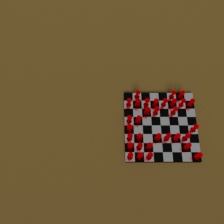}
    \includegraphics[width=\thumbwidth]{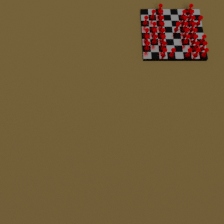}
    \label{fig:symtrans}
  }

  \subfloat[Random camera position on sphere.]{
    \includegraphics[width=\thumbwidth]{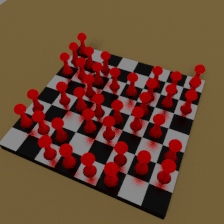}
    \includegraphics[width=\thumbwidth]{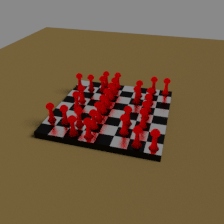}
    \includegraphics[width=\thumbwidth]{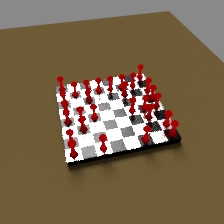}
    \hspace{\imgspace}
    \includegraphics[width=\thumbwidth]{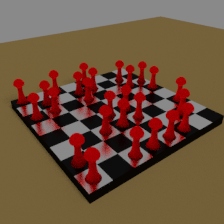}
    \includegraphics[width=\thumbwidth]{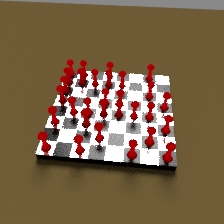}
    \includegraphics[width=\thumbwidth]{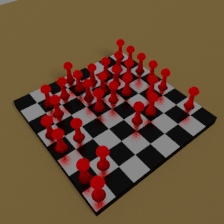}
    \label{fig:symrot}
  }
  \caption{Different, tested variations of the symmetry--task. The
    left group of images are symmetric. The right group are not
    symmetric and have ten pawns that are out of place.}
\end{figure*}

Example images for both of these tasks can be seen in
\autoref{fig:intro}. The difficulty of both tasks can be controlled in
multiple different ways:
\begin{enumerate}
\item The number of pawns, breaking the symmetry/identity, can be
  adjusted. It should be evident that detecting a single, out of place
  pawn is more difficult than detecting ten pawns that are out of
  place.
\item The task can be made more challenging by increasing the visual
  variability of the presented images. We are using three different
  levels of variability:
  \begin{enumerate}
  \item For the lowest amount of variability, the camera position as well
    as the board positions are fixed. See \autoref{fig:simfix} and
    \autoref{fig:symfix} for example images.
  \item For more variability, the camera is randomly moved on a plane,
    resulting in different board positions for each of the images.
    Still, in the identity--task, the relative position of the two
    boards stays the same. See \autoref{fig:simtrans} and
    \autoref{fig:symtrans} for example images.
  \item The highest variability is achieved by randomly positioning
    the camera on a sphere, with a random radius, around the
    checkerboard. This results in different view points, as well as
    different sizes of the boards. See \autoref{fig:simrot} and
    \autoref{fig:symrot} for example images.
  \end{enumerate}
\item For the symmetry--task, the two checkerboards can be arranged
  randomly. See \autoref{fig:simrand} for example images.
\end{enumerate}
For each of these variability--schemes, we test versions of the
dataset with one, five, and ten pawns that are out of place.

The images for the dataset were generated by an automated procedure,
using the 3D modeling software Blender \cite{blender}. Thus, an
arbitrary amount of training-- and testing--images can be produced
very quickly in arbitrary resolutions. This also opens the door to
varying lighting conditions, added clutter, additional chess pieces,
\dots, to make the dataset more challenging. The scripts to generate
the dataset can be found
online\footnote{\url{https://github.com/Paethon/chess_image_dataset}}.

\section{Experiments}
To evaluate the dataset, we generated 20000 training images as well as
1000 testing images for each of the tasks and difficulty levels. A
resolution of $224 \times 224$ pixels was used. We tested the dataset
on the popular network architectures
Alexnet~\cite{krizhevsky2012imagenet}, VGG16~\cite{simonyan2014very},
and GoogLeNet~\cite{szegedy2015going}.

For AlexNet and GoogLeNet, the standard implementations provided with
the nVidia DIGITS~\cite{digits} deep learning framework version 5.0.0
using Caffe~\cite{jia2014caffe} version 0.15.13 as a back end were
used. Since we were not able to train VGG16 on the presented dataset
from scratch, we used the predefined network, pre--trained on
ImageNet, from the DIGITS model store. All networks were trained using
ADAM, with a base learning rate of $1 \times 10^{-5}$ for 120 epochs.
The training was manually stopped in cases where further improvement
was not to be expected (e.g. perfect accuracy was already achieved).

For each task, the networks were trained in order of increasing
difficulty, and the learned weights were used as initialization for
the next, more difficult, task. A network was, for example, trained on
the symmetry--task with a fixed camera position and ten out of place
pawns. After successful training of this network, the weights were
used to initialize the network to be trained on the same task with
five out of place pawns. A variation of this approach was presented by
Bengio \etal \cite{bengio2009curriculum} under the name of
\emph{curriculum learning}. This approach was absolutely critical for
training some of the more difficult variations of the dataset. We
were, for example, not able to achieve a classification accuracy above
chance with GoogLeNet on the identity--task with random board
positions (\autoref{fig:simrand}) without this curriculum learning
approach, despite the fact that we could reach a good classification
accuracy of 0.86 using curriculum learning.

\begin{table*}[t]
  \centering
  \caption{Highest achieved accuracies on the proposed dataset by the
    tested CNN architectures.}
  \begin{tabular}{lccccc}
    \toprule
    Task & AlexNet & VGG16 & GoogLeNet\\
    \midrule
    \midrule
    \textbf{identity} &  &  & \\
    fixed position, 10 diff (Fig.\ref{fig:simfix}) & 1.00 & 1.00 & 0.99\\
    fixed position, 5 diff & 1.00 & 0.99 & 0.97\\
    fixed position, 1 diff & 0.99 & 1.00 & 1.00\\
    \midrule
    camera translation, 10 diff (Fig.\ref{fig:simtrans}) & 0.99 & 0.99 & 0.99\\
    camera translation, 5 diff & 0.98 & 0.99 & 0.98\\
    camera translation, 1 diff & 0.90 & 0.98 & 0.96\\
    \midrule
    random board placement, 10 diff (Fig.\ref{fig:simrand}) & 0.80 & 0.89 & 0.95\\
    random board placement, 5 diff & 0.73 & 0.88 & 0.94\\
    random board placement, 1 diff & 0.54 & 0.69 & 0.86\\
    \midrule
    camera rotation, 10 diff (Fig.\ref{fig:simrot}) & 0.54 & 0.64 & 0.55\\
    camera rotation, 5 diff & 0.52 & 0.63 & 0.53\\
    camera rotation, 1 diff & 0.51 & 0.54 & 0.50\\
    \midrule
    \midrule
    \textbf{symmetry} &  &  & \\
    fixed position, 10 diff (Fig.\ref{fig:symfix}) & 1.00 & 1.00 & 1.00\\
    fixed position, 5 diff & 1.00 & 1.00 & 1.00\\
    fixed position, 1 diff & 0.99 & 1.00 & 1.00\\
    \midrule
    camera translation, 10 diff (Fig.\ref{fig:symtrans}) & 0.99 & 1.00 & 1.00\\
    camera translation, 5 diff & 0.98 & 0.99 & 0.98\\
    camera translation, 1 diff & 0.85 & 0.99 & 0.92\\
    \midrule
    camera rotation, 10 diff (Fig.\ref{fig:symrot}) & 0.75 & 0.85 & 0.79\\
    camera rotation, 5 diff & 0.59 & 0.80 & 0.78\\
    camera rotation, 1 diff & 0.52 & 0.59 & 0.63\\
    \bottomrule
  \end{tabular}
  \label{tab:results}
\end{table*}

During training, each network was evaluated on the testing set after
each epoch, and an accuracy measure was recorded. Accuracy is defined
as $\frac{|s_c|}{|s|}$ where $|s|$ is the number of tested samples
(\ie the number of images to be classified) and $|s_c|$ is the number
of correctly classified samples. Since we have two possible classes
for all our experiments, a purely random classifier would achieve an
accuracy of $\approx 0.5$. For each network and task, we report the
highest achieved accuracy for all of the evaluations, after each of
the 120 training epochs. We thus expect even a random classifier to
get a maximum accuracy above 0.5. If we assume an equal probability
for both classes, 1000 samples classified per test, and 120 tests, we
expect a purely random classifier to achieve a mean maximum accuracy
over all 120 evaluations of $\approx 0.54$, with a standard deviation
of $\approx 6.6 \times 10^{-3}$. These values were determined using
simulation.
\subsection{Discussion}

\autoref{tab:results} shows the highest achieved accuracy during
training. The identity--task with fixed camera position and camera
translation (\autoref{fig:simfix}) was solved almost perfectly by all
tested network architectures. This is not very surprising, since the
same checker board positions will always be at the same pixel
positions. Thus, the networks can learn a very direct mapping, to
check for identity and symmetry.

Somewhat more surprising is the almost perfect performance of all
three networks on the dataset variation with random camera translation
(\autoref{fig:simtrans}). Especially, since the translation of the
camera also imparts perspective effects on the images (\ie if the
checkerboard is rendered at the top of the image, it is smaller in
comparison to being rendered at the bottom). Still, the relative
position of all the checkerboard positions is constant in all the
images, up to some scaling factor. This might explain the overall good
performance of the networks. AlexNet does perform somewhat worse with
only one pawn out of place, but it still reaches a good accuracy of
0.90.

Random board placement and fixed camera angle (\autoref{fig:simrand})
is interesting, since the tested architectures perform very
differently on this task. GoogLeNet performs very well, even solving
one pawn out of place well above chance. AlexNet performs much worse,
and does not solve the one pawn out of place variant at all. VGG16
lies somewhere in the middle. The less than perfect performance is
interesting, since human subjects would very likely not consider this
task more difficult than the variations with fixed camera position, or
camera translation. It could be the case, that features have to be
integrated on a more global scope than in the other tasks, which leads
to diminished performance.

The variant with camera rotation (\autoref{fig:simrot}) was not solved
convincingly by any of the architectures. VGG16 performs slightly
better than chance, with an accuracy of 0.64 and 0.63 for ten and five
out of place pawns respectively, but it also completely fails with
only one out of place pawn. The images that VGG16 can correctly
classify predominantly show the checkerboard in a very favorably
position (\ie top-down with little rotation). AlexNet and GoogLeNet
seem to be confused enough by the rest of the training set so that
they are not even able to classify these easier images.

The symmetry--task seems to be easier for the networks in general.
This likely has two reasons. On one hand, only 64 board positions have
to be compared in comparison with 128 positions for the
identity--task. On the other hand, the positions to be compared are
also spatially closer, especially for the more difficult variations of
the dataset.

The variation with fixed camera and board position
(\autoref{fig:symfix}) is solved perfectly by all the networks. Added
camera translation (\autoref{fig:symtrans}) shows a similar pattern to
what we have seen for the identity--task. All networks solve this
problem more or less perfectly, except for AlexNet, which is only able
to achieve an accuracy of 0.85 for one out of place pawn. This
suggests that there seems to be a general flaw in the AlexNet
architecture for these kinds of problems.

Adding camera rotation (\autoref{fig:symrot}) leads to more variable
results. None of the networks perform perfectly, but all of them
perform significantly above chance for the variation with ten out of
place pawns. VGG16 and GoogLenet even perform slightly above chance
for one out of place pawn.

The experiments reveal a few variations of the dataset that seem to be
interesting for further research:
\begin{enumerate}
\item Symmetry--task with camera rotation: This variant seems to be at
  the border of being solvable by current architectures and the
  difficulty scales well with the number of out of place pawns.
\item Identity--task with random board placement: The network
  architecture seems to be especially relevant for this task.
\item Identity--task with camera rotation: None of the networks were
  able to solve any of the variants of this task convincingly, but the
  fact that VGG16 does perform slightly above chance indicates that it
  might be possible to create a network architecture that performs
  much better.
\end{enumerate}

It would be interesting to evaluate these variations of the dataset on
additional network architectures, and to analyze how human subjects
solve problems of this kind. Our hypothesis is that such problems are
generally not solved in a pure feed forward manner by humans, and some
attentional mechanisms and iterative processing of the images are
required. Attention is defined by the Encyclopedia Britannica as ``the
concentration of awareness on some phenomenon to the exclusion of
other stimuli''. Since brains do have capacity limitations, it is
impossible to process all visual information at any given time, as
shown by Tsotsos \cite{tsotsos1990analyzing}. Therefore, an
attentional mechanism has to assign the available resources to task
relevant stimuli. We hypothesize that pawn positions are compared not
as a whole, but by an iterative switching of attention between smaller
areas of the board or boards. To substantiate this hypothesis, we
propose to test the classification accuracy and classification speed
of human subjects on the same dataset, while also collecting eye
tracking data, to get a rough estimate of shifting attention.
Processing of the images in this way would hint at the possibility
that attention and iterative processes might be more efficient at, or
even necessary, for solving the problem classes presented in our
dataset.

It would also be interesting to see whether the time humans need to
correctly classify an image correlates with the classification
performance of a CNN. A human might for example need less time to
classify a pawn arrangement if a pawn is misplaced in one of the
corners.

It would also be interesting to see whether current CNN architectures
that already incorporate some form of attention, as well as a form of
iterative processing of images, would perform better on the dataset
than the already tested standard architectures. Sermanet \etal
\cite{sermanet2014attention} have shown that incorporating attention
and iterative refinement of class predictions can improve the
performance of CNNs.

\section{Conclusion}
We presented a novel image classification dataset that should be
trivial to classify for humans. Nonetheless, certain variations of it
are poorly classified by the tested CNN architectures AlexNet, VGG16,
and GoogLeNet. We identified three variations of the dataset that
might be interesting for further research. Detecting symmetry of pawn
positions of a checkerboard, together with camera rotation, is
interesting, since it seems to be on the border of what current CNN
architectures can solve. Depending on the number of pawns that break
the symmetry, it can, or can not be solved. Detecting identity of pawn
positions on two randomly positioned checkerboards, with fixed camera
position, is the second interesting variation of the dataset. From our
perspective, it seems like it should be an easy task for human
subjects, but the tested architectures showed highly variable
performance. Third, the identity--task, with camera rotation, was not
convincingly solved by any of the architectures. We therefore proposed
to do additional tests on these specific variations of the dataset. In
addition, experiments involving human subjects might be interesting to
determine under which circumstances and by which processes humans are
able to classify this dataset. Our hypothesis is that humans use some
form of attentional mechanism and iterative processing to solve
problems of this kind. We further hypothesize that such an approach is
therefore more efficient for the given task at hand, and incorporating
these principles might benefit machine learning methods.

We want to thank the reviewers for the helpful comments and proposing
further research.

{\small
  \bibliographystyle{ieee}
  \bibliography{egbib}

\begin{thebibliography}{10}\itemsep=-1pt

\bibitem{bengio2009curriculum}
Y.~Bengio, J.~Louradour, R.~Collobert, and J.~Weston.
\newblock Curriculum learning.
\newblock In {\em Proceedings of the 26th annual international conference on
  machine learning}, pages 41--48. ACM, 2009.

\bibitem{blender}
{Blender Online Community}.
\newblock {\em Blender - a 3D modelling and rendering package}.
\newblock Blender Foundation, Blender Institute, Amsterdam, 2017.

\bibitem{bongard1970}
M.~M. Bongard.
\newblock {\em Pattern Recognition}.
\newblock Spartan Books, 1970.

\bibitem{dodge2017study}
S.~Dodge and L.~Karam.
\newblock A study and comparison of human and deep learning recognition
  performance under visual distortions.
\newblock {\em arXiv preprint arXiv:1705.02498}, 2017.

\bibitem{Everingham10}
M.~Everingham, L.~Van~Gool, C.~K.~I. Williams, J.~Winn, and A.~Zisserman.
\newblock The pascal visual object classes (voc) challenge.
\newblock {\em International Journal of Computer Vision}, 88(2):303--338, June
  2010.

\bibitem{fleuret2011comparing}
F.~Fleuret, T.~Li, C.~Dubout, E.~K. Wampler, S.~Yantis, and D.~Geman.
\newblock Comparing machines and humans on a visual categorization test.
\newblock {\em Proceedings of the National Academy of Sciences},
  108(43):17621--17625, 2011.

\bibitem{he2015delving}
K.~He, X.~Zhang, S.~Ren, and J.~Sun.
\newblock Delving deep into rectifiers: Surpassing human-level performance on
  imagenet classification.
\newblock In {\em Proceedings of the IEEE international conference on computer
  vision}, pages 1026--1034, 2015.

\bibitem{jia2014caffe}
Y.~Jia, E.~Shelhamer, J.~Donahue, S.~Karayev, J.~Long, R.~Girshick,
  S.~Guadarrama, and T.~Darrell.
\newblock Caffe: Convolutional architecture for fast feature embedding.
\newblock {\em arXiv preprint arXiv:1408.5093}, 2014.

\bibitem{krizhevsky2009learning}
A.~Krizhevsky and G.~Hinton.
\newblock Learning multiple layers of features from tiny images.
\newblock 2009.

\bibitem{krizhevsky2012imagenet}
A.~Krizhevsky, I.~Sutskever, and G.~E. Hinton.
\newblock Imagenet classification with deep convolutional neural networks.
\newblock In {\em Advances in neural information processing systems}, pages
  1097--1105, 2012.

\bibitem{lecun1998gradient}
Y.~LeCun, L.~Bottou, Y.~Bengio, and P.~Haffner.
\newblock Gradient-based learning applied to document recognition.
\newblock {\em Proceedings of the IEEE}, 86(11):2278--2324, 1998.

\bibitem{digits}
nVidia.
\newblock {DIGITS} interactive deep learning gpu training system.
\newblock \url{https://developer.nvidia.com/digits}.
\newblock Accessed: 2017-07-29.

\bibitem{russakovsky2015imagenet}
O.~Russakovsky, J.~Deng, H.~Su, J.~Krause, S.~Satheesh, S.~Ma, Z.~Huang,
  A.~Karpathy, A.~Khosla, M.~Bernstein, et~al.
\newblock Imagenet large scale visual recognition challenge.
\newblock {\em International Journal of Computer Vision}, 115(3):211--252,
  2015.

\bibitem{sermanet2014attention}
P.~Sermanet, A.~Frome, and E.~Real.
\newblock Attention for fine-grained categorization.
\newblock {\em arXiv preprint arXiv:1412.7054}, 2014.

\bibitem{simonyan2014very}
K.~Simonyan and A.~Zisserman.
\newblock Very deep convolutional networks for large-scale image recognition.
\newblock {\em arXiv preprint arXiv:1409.1556}, 2014.

\bibitem{stabinger201625}
S.~Stabinger, A.~Rodr{\'\i}guez-S{\'a}nchez, and J.~Piater.
\newblock 25 years of cnns: Can we compare to human abstraction capabilities?
\newblock In {\em International Conference on Artificial Neural Networks},
  pages 380--387. Springer, 2016.

\bibitem{stabinger2016learning}
S.~Stabinger, A.~Rodr{\'\i}guez-S{\'a}nchez, and J.~Piater.
\newblock Learning abstract classes using deep learning.
\newblock In {\em proceedings of the 9th EAI International Conference on
  Bio-inspired Information and Communications Technologies (formerly BIONETICS)
  on 9th EAI International Conference on Bio-inspired Information and
  Communications Technologies (formerly BIONETICS)}, pages 524--528. ICST
  (Institute for Computer Sciences, Social-Informatics and Telecommunications
  Engineering), 2016.

\bibitem{szegedy2015going}
C.~Szegedy, W.~Liu, Y.~Jia, P.~Sermanet, S.~Reed, D.~Anguelov, D.~Erhan,
  V.~Vanhoucke, and A.~Rabinovich.
\newblock Going deeper with convolutions.
\newblock In {\em Proceedings of the IEEE conference on computer vision and
  pattern recognition}, pages 1--9, 2015.

\bibitem{Todorovic:2008}
D.~Todorovic.
\newblock {G}estalt principles.
\newblock {\em Scholarpedia}, 3(12):5345, 2008.
\newblock revision \#91314.

\bibitem{tsotsos1990analyzing}
J.~K. Tsotsos.
\newblock Analyzing vision at the complexity level.
\newblock {\em Behavioral and brain sciences}, 13(3):423--445, 1990.

\end{thebibliography}
}

\end{document}